 \documentclass[final,5p,times,twocolumn,authoryear]{elsarticle}

\usepackage{amssymb}
\usepackage{amsmath}
\usepackage{lipsum}
\usepackage{float}
\usepackage{subcaption}
\usepackage{algorithm}
\usepackage[noend]{algorithmic}

\journal{Computers and Security}

\begin{document}

\begin{frontmatter}



\title{PATE-TripleGAN: Privacy-Preserving Image Synthesis with Gaussian Differential Privacy}


\author[first]{Jiang Zepng}
\author[second]{Ni Weiwei}
\author[second]{Zhang Yifan}

\affiliation[first]{
    Department={College of Software Engineering},
    Organization={South-East University},
    city={NanJing},
    postcode={211189}, 
    country={China}
}
\affiliation[second]{
    Department={School of Computer Science and Engineering},
    Organization={South-East University},
    city={NanJing},
    postcode={211189}, 
    country={China}
}

\begin{abstract}
Conditional Generative Adversarial Networks (CGANs) exhibit significant potential in supervised learning model training by virtue of their ability to generate realistic labeled images. However, numerous studies have indicated the privacy leakage risk in CGANs models. The solution DPCGAN, incorporating the differential privacy framework, faces challenges such as heavy reliance on labeled data for model training and potential disruptions to original gradient information due to excessive gradient clipping, making it difficult to ensure model accuracy. To address these challenges, we present a privacy-preserving training framework called PATE-TripleGAN. This framework incorporates a classifier to pre-classify unlabeled data, establishing a three-party min-max game to reduce dependence on labeled data. Furthermore, we present a hybrid gradient desensitization algorithm based on the Private Aggregation of Teacher Ensembles (PATE) framework and Differential Private Stochastic Gradient Descent (DPSGD) method. This algorithm allows the model to retain gradient information more effectively while ensuring privacy protection, thereby enhancing the model's utility. Privacy analysis and extensive experiments affirm that the PATE-TripleGAN model can generate a higher quality labeled image dataset while ensuring the privacy of the training data.
\end{abstract}




\end{frontmatter}




\section{Introduction}
\label{introduction}
Nowadays data synthesis plays an increasingly crucial role in enhancing the generalization ability of machine learning models by providing vast training samples \citep{lu2023machine}, especially in the domain of supervised learning. The widely adopted Generative Adversarial Networks (GANs) algorithm proposed by \cite{goodfellow2014generative}, introduced for efficient feature capturing through a two-party game between generator and discriminator, has become the mainstream paradigm for data synthesis. Building upon GANs, Conditional GANs (CGANs) further enhance the synthesis process by introducing conditional information to the generator, enabling the generation of samples that adhere to specific constraints \citep{mirza2014conditional}. This advancement has yielded significant achievements in various domains, including image generation \citep{ma2021ml,hu2024crd} and style transfer \citep{liu2021sccgan,qin2022style}.

However, as the importance of data privacy grows, concerns about privacy leakage during CGANs training have surfaced. The GANs-based models, while generating new data samples, may inadvertently reveal sensitive information from the underlying datasets \citep{liu2020privacy}. Moreover, attacks such as membership inference \citep{shokri2017membership} and model inversion\citep{fredrikson2015model} pose risks of reconstructing partial data and stealing sensitive privacy information. Hence, safeguarding the privacy of training data has become a pressing need in the field of data synthesis while aiming to generate high-quality samples.

In response to potential privacy attacks, \cite{torkzadehmahani2019dp} proposed the Differential Private Conditional GANs (DPCGANs) model under the framework of differential privacy \citep{dwork2006differential}. By individually clipping the loss gradients of real and synthetic data, DPCGAN achieves precise control over sensitivity. Additionally, it employs a gradient desensitization technique based on Rényi-DP \citep{mironov2017renyi} to precisely track privacy budget consumption, ensuring the generation of labeled datasets while preserving data privacy. Nevertheless, DPCGAN relies heavily on extensive fully labeled data. The inability to leverage unlabeled real data for supplementary training significantly diminishes the model's applicability. In addition, the clipping of the gradient needs to be finely adjusted according to the privacy budget. Excessive clipping will destroy the original gradient information, leading to a decrease in the accuracy of the model.

Triple-GAN \citep{li2017triple} can effectively alleviate the heavy reliance of CGANs on labeled data by introducing a classifier into the GANs framework. However, under the Triple-GAN framework, the loss function of the discriminator covers more information. Directly applying the differentially private stochastic gradient descent (DPSGD) algorithm\citep{abadi2016deep} to clip and noisy the discriminator's gradients will seriously damage the original gradient information, thus potentially compromise the convergence of Triple-GAN.

To address these challenges, we propose a novel privacy protection training framework—PATE-TripleGAN. This framework allows for training on unlabeled data while ensuring privacy, producing high-quality labeled images. The core idea of PATE-TripleGAN is to introduce a classifier to preclassify unlabeled data, transforming the training of the model from supervised learning of CGAN to semi supervised learning. For the synthesized dataset sampled from the classifier and generator, algorithm based on DPSGD\citep{abadi2016deep} and noisy aggregation based on the PATE mechanism \citep{papernot2016semi} are used to iteratively train the discriminator network by generating their respective desensitization gradients, thereby retaining gradient information more effectively and achieving a balance between privacy protection and model utility.

The following is a summary of this article's work and contributions:
\begin{itemize}
\item We propose a novel privacy protection framework, called PATE-TripleGAN, by establishing a three-party min-max game involving a generator, a classifier, and a discriminator based on the PATE mechanism. It adeptly navigates the challenge of heavy reliance on labeled data, achieving a balance between privacy protection and the quality of generated images.
\item We present a hybrid gradient desensitization algorithm. It employs different gradient desensitization methodologies predicated on dissimilarities in features within synthetic datasets, thus effectively retaining more original gradient information.
\item Rigorous benchmark testing on MNIST\citep{lecun1998mnist} and Fashion-MNIST\citep{xiao2017fashion} datasets underscores the effectiveness of our solution. Both privacy analysis and experimental results demonstrate that our model can generate a higher quality labeled image dataset while ensuring the data privacy.
\end{itemize}

The structure of this paper is outlined as follows: Section 2 reviews relevant achievements in existing research literature. Section 3 provides details and background knowledge of reference models. Section 4 elucidates the implementation process and algorithms of our model. Section 5 presents the comprehensive evaluation results from experiments. Section 6 summarizes and discusses the effectiveness of the model.

\section{Related Work}

This section provides an overview of the relevant literature and work that significantly to the research presented in this paper.

\subsection{Generative Adversarial Networks}
The introduction of the Generative Adversarial Networks framework marked a pivotal advancement in data synthesis. GANs operate by training discriminators and generators alternately and iteratively, enabling the generator to produce highly realistic synthetic data \citep{goodfellow2014generative}. Expanding on the GAN framework, conditional Generative Adversarial Networks were proposed \citep{mirza2014conditional}. CGANs incorporate conditional constraints, such as data labels, and random noise as inputs to the generator. The integration of conditional elements in CGANs enhances the model's ability to generate samples meeting specific requirements, playing a crucial role for supervised learning. However, CGANs’ training heavily relies on labeled data, limiting the applicability of models. To overcome this constraint, Triple-GAN introduces a classifier to pretreat the unlabeled data. By incorporating a three-party min-max game, the training of the model is transformed into a semi-supervised learning process \citep{li2017triple}, relieving the dependence of model training on fully labeled data.

While GANs have achieved remarkable successes in supervised learning, the looming threat of potential privacy attacks poses a serious risk to the data security of GANs models. Addressing the pressing need to generate high-quality data while ensuring privacy becomes imperative.

\subsection{Differential Privacy and GANs}
Researchers propose integrating privacy protection frameworks with deep learning models to address potential privacy threats\citep{abadi2016deep}. DPGAN\citep{xie2018differentially}, based upon differential privacy, introduces differential noise to the discriminator's gradient, achieving privacy protection for the generator through the post-processing nature \citep{dwork2014algorithmic} of differential privacy. Similarly, the Gs-WGAN model \citep{chen2020gs} demonstrates the feasibility of implementing a similar gradient desensitization approach to the generator. Furthermore, \cite{jordon2018pate} suggests incorporating the PATE mechanism into the discriminator. This involves training a batch of teacher models on a sensitive dataset and then using synthetic data sampled from the generator for knowledge transfer based on PATE to train a student discriminator. 

However, these models overlook the generation of labeled data, making the synthetic data unsuitable for direct use in supervised learning. To address this issue, \cite{torkzadehmahani2019dp} proposed the DPCGAN model, transforming the generator's input into a combination of random noise and constraint conditions. By clipping the loss functions separately for real and synthetic samples, DPCGAN achieves precise control over sensitivity. The model also employs a gradient desensitization method based on the Renyi-DP \citep{mironov2017renyi}, enabling accurate tracking of privacy budgets. Nonetheless, DPCGAN cannot overcome the dilemma of heavy reliance on labeled data. Additionally, excessive clipping and high-dimensional noise may destroy the original gradient information under tight privacy budgets, leading to a decrease in model accuracy and convergence difficulties.

Even when integrating Triple-GAN with a differential privacy framework, numerous challenges arise. Firstly, we find the pre-classification of unlabeled data by Triple-GAN will introduce disparate sensitivities between labeled and unlabeled datasets, demanding separate treatment. Furthermore, the inclusion of a classifier amplifies the complexity of the discriminator's loss function by incorporating additional information. Hasty measures such as gradient clipping and noise addition can significantly distort the original gradient information. These challenges make it difficult to directly integrate well-established differential privacy frameworks with Triple-GAN, posing obstacles to ensuring the privacy security of Triple-GAN models.

\section{Preliminaries}
\subsection{Gaussian Differential Privacy}
Differential privacy sets a rigorous standard for privacy protection in machine learning algorithms. It aims to ensure that, for any query mechanism $M$, the probability of $M$ producing the same result on neighboring datasets (differing by only one data point) is close to 1 \citep{dwork2006differential}. 

\textbf{Definition 1.} (Differential Privacy)
A random mechanism $M$ satisfies $(\epsilon,\delta)-DP\;(\epsilon >0, \delta \geq 0)$, if for any $S_m \in O$ and for any neighboring datasets $D,\;D'$:
\begin{equation}
    P_r[M(D)\in S_m] \leq e^\epsilon \times P_r[M(D')\in S_m]+\delta
\end{equation}

The newly proposed Gaussian-DP framework \citep{dong2022gaussian}, based on hypothesis testing, offers an equivalent definition of differential privacy.

\textbf{Definition 2.} (Gaussian-DP)
A random mechanism $M$ satisfies $\mu-GDP \;(\mu > 0)$ if for any neighboring datasets $D, \;D'$:
\begin{equation}
    T(M(S), M(S'))\geq G_\mu, \;with\; G_\mu =T(\mathcal{N}(0,1),\mathcal{N}(\mu,1))
\end{equation}

The term $f = T(M(S), M(S'))$ signifies that a mechanism $M$ is $f-DP$ if distinguishing any two neighboring datasets is no easier than distinguishing between $S$ and $S'$ \citep{dong2022gaussian}. In Gaussian-DP, the composite of  privacy mechanisms is denoted by the tensor product $\otimes$, which is both commutative and associative. This means that for $f = T(P, Q)$ and $g = T(P', Q')$:
\begin{equation}
    f \otimes g = g \otimes f := T(P \times P', Q \times Q')
\end{equation}

Specifically, the gaussian mechanism with sensitivity $\Delta$ f and standard deviation $\sigma$ adheres to $\frac{\Delta f}{\sigma}-GDP$. When dealing with the composite of  pure Gaussian mechanisms, the expressions in Gaussian-DP become remarkably straightforward.

\textbf{Lemma 1.} For n Gaussian mechanisms with parameters $\mu_1,\; \mu_2,\;...,\;\mu_n$, their composition mechanism still satisfies Gaussian-DP:
\begin{equation}
    G_{\mu_1} \otimes G_{\mu_2} \;...\; G_{\mu_n } = G_\mu, \; with \; \mu = \sqrt{\mu_1^2+\mu_2^2+...+\mu_n^2} 
\end{equation}

The Gaussian-DP's composition theorem, based on hypothesis testing without intricate scaling operations, offers more stringent privacy boundaries compared to the strong composition theorem \citep{kairouz2015composition} and Moments Accountant \citep{abadi2016deep}. Additionally, Gaussian-DP can achieve equivalent transformation to classical $(\epsilon,\delta)-DP$ through the following equation:
\begin{equation}
    \delta(\epsilon) = \Phi (\;-\frac{\epsilon}{\mu}+\frac{\mu}{2}) - e^{\epsilon}\;\Phi(\;-\frac{\epsilon}{\mu}-\frac{\mu}{2})
\end{equation}

We opted for the Gaussian-DP as the privacy framework for our model due to its advantageous compositional properties. In the experimental section, we transformed it into a unified format of $(\epsilon,\delta)-DP$ for comparative analysis.

\subsection{Triple-GAN}
Triple-GAN establishes a three-player game involving a generator, a classifier, and a discriminator. The generator produces fake-data for a real label, while the classifier generates fake-labels for real data. The discriminator distinguishes between the authenticity of these ``data-label" pairs, updating the model through gradient back-propagation during iterative training. The revised discriminator loss function for Triple-GAN is:
\begin{equation}
\begin{split}
\label{Triple-GAN}
     \mathop{min}\limits_{C,G} \;\mathop{max}\limits_{D} \;E_{P}[\log(D(x,y)]+\alpha E_{P_c}[\log(1-D(x,y))] \\
    + \;(1-\alpha)E_{P_g}[\log(1-D(G(y,z),y))],\;\alpha\in(0,1)
\end{split}
\end{equation}

To ensure model convergence, Triple-GAN introduces KL divergence $D_{KL}(P(x,y)||P_c(x,y))$ between the classifier and real data distribution, and $D_{KL}(P_g(x,y)||P_c(x,y)$ between the generator and the classifier. These constraints prevent a competitive relationship between the classifier and the generator.

We employ Triple-GAN as the primary framework instead of CGAN, integrating a classifier to address the model's heavy reliance on labeled data during training. It's essential to note that this framework will introduce new sensitivity from unlabeled dataset, while also amplifying the information from the discriminator. Directly clipping and noising the gradient may significantly destroy the original gradient information.

\subsection{PATE Mechanism}
The PATE mechanism protects privacy by sharing noisy information from teacher models to a student model. It works on the idea that if two classifiers trained on different sets give the same output for a given input, it won't disclose the privacy of any individual training sample\citep{papernot2016semi}.

During the training phase, the PATE mechanism partitions the sensitive dataset into $k$ disjoint subsets, independently training a teacher classifier on each subset. Subsequently, the student classifier obtains non-private data from a public source and aggregates the votes of teacher models through a noisy voting process. Each teacher model provides an independent classification result:
\begin{equation}
    N_i(X)= |\; j:j\in [1,...,k], T_j(X)=Label_i\; |, \; i\in[1,...,n]
\end{equation}

The aggregation mechanism outputs the category with the maximum noisy vote as the final classification result for the query:
\begin{equation}
    PATE_\lambda(X=\mathop{argmax}_{i\in [m]}(N_i(X)+Lap(\lambda))
\end{equation}

The student classifier learns from noisy voting results. The PATE mechanism only reveals the student model, safeguarding model privacy.

Another method, called GNMax aggregation, employs the Gaussian mechanism's sensitivity to outliers to design threshold conditions for combining teacher models' responses to queries\citep{papernot2018scalable}. With threshold filtering, GNMax aggregation can significantly improve the efficiency of privacy budget utilization.

We choose the PATE mechanism to desensitize the private gradient of the generator part in the discriminator, which can preserve more original information compared to clipping the entire gradient.

\begin{figure*}[t]
	\centering 
	\includegraphics[width=0.8\textwidth]{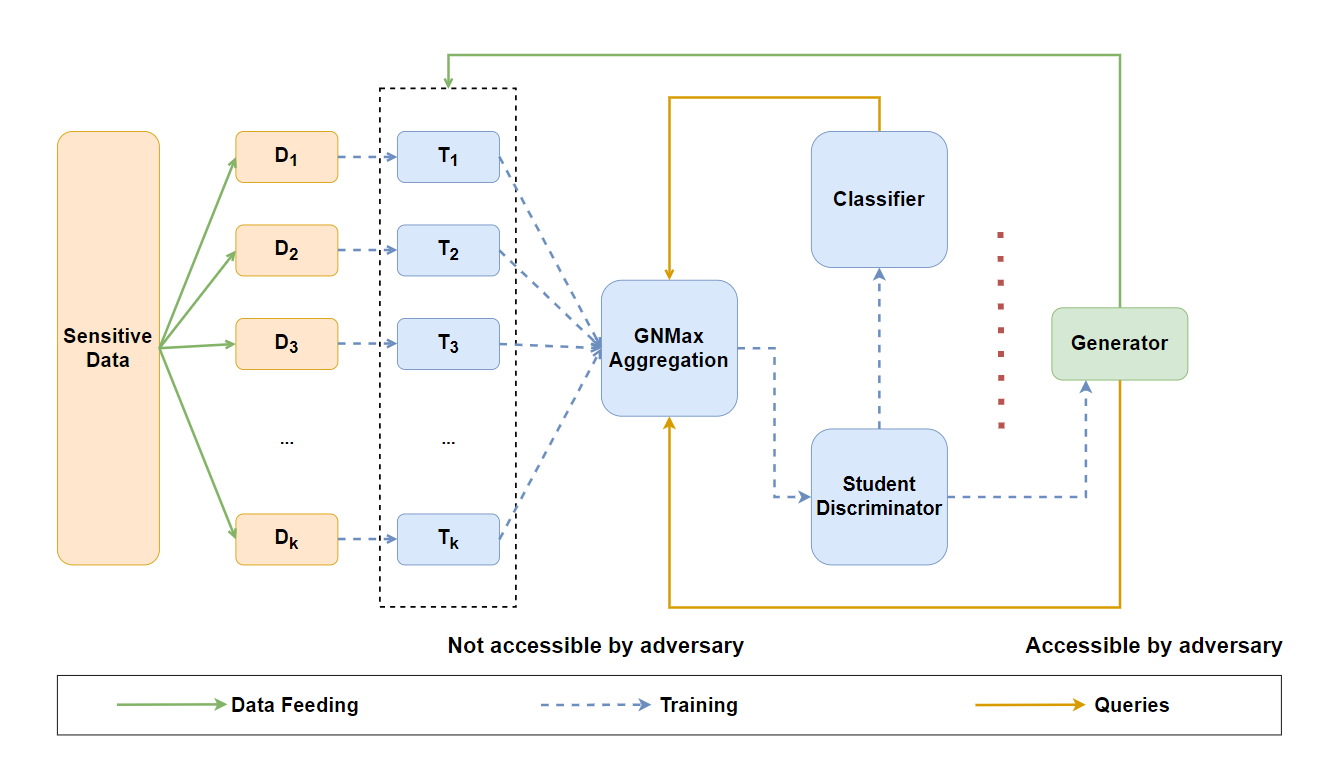}	
	\caption{The framework of PATE-TripleGAN.} 
	\label{framework}%
\end{figure*}

\section{Methodology}
In this section, we will present the specific details of the PATE-TripleGAN model. For objects in the form of ``data-label", privacy is defined by: ($I$) the inherent feature privacy of the data itself and ($II$) the associate privacy of the correspondence between data and labels (Individual label information itself is not considered private). Correspondingly, the sensitive dataset consists of two parts: ($I$) the labeled dataset $S_l$ and ($II$) the unlabeled data set $S_d$.

PATE-TripleGAN establishes a three-party min-max game, including a generator, a classifier, and a discriminator based on the PATE mechanism, as illustrated in Figure \ref{framework}. To alleviate the model's heavy reliance on labeled data during training, we introduce a classifier to pre-classify $S_d$, transforming the model training from CGAN's supervised learning to semi-supervised learning. For the synthetic datasets sampled from the classifier and generator, we employ DPSGD algorithm and PATE-based noisy aggregation to generate their respective privacy-preserving gradients, iteratively training the discriminator network. The privacy protection for the generator is achieved through the post-processing property\citep{dwork2014algorithmic}.

Compared to the classical approach of adding differential noise uniformly to clipped gradients for privacy protection, PATE-TripleGAN can more effectively retain information from the original gradients, enhancing the model's utility and convergence. Theoretical analysis demonstrates that PATE-TripleGAN can achieve rigorous protection in both ``data feature privacy" and ``data-label correspondence privacy”.

We mainly discuss our work from the following aspects: ``Teacher Models", ``Aggregation Mechanism",``GANs Framework" and ``Privacy Analysis".

\subsection{Teacher Models}
In PATE-TripleGAN, the teacher model functions as a binary classifier trained to discern the authenticity of ``data-label" pairs. It learns not only from the partitioned sensitive dataset but also from the sampled fake ``data-label" pairs from the generator to enhance its discriminative capability.

We partition the labeled sensitive dataset \(S_l\) into \(k\) disjoint subsets, denoted as \(D_1, D_2, \ldots, D_k\), each containing \(|D|/k\) data instances. Subsequently, each subset corresponds to a teacher model (\(T_1, T_2, \ldots, T_k\)). For true sensitive ``data-label" pairs, we assign the result 1; for fake ``data-label" pairs sampled from the generator, we assign the result 0. 

When training teacher models, it's crucial to avoid incorporating the classifier's data as part of the training set. This is because, for the classifier, the generated ``data" inherently involves the feature privacy of $S_d$, falling under privacy requirement ($I$). Moreover, introducing the data generated by the classifier directly into the teacher model will also disrupt the sensitivity of the original partition.

The loss function for the \(i\)-th teacher is defined as:
\begin{equation}
    L_{T_i} = -[\sum_{(x_d,y_d)\in D_i}logT_i(x_d,y_d)+\sum_{(x_t,y_t)\in G}log(1-T_i(x_t,y_t) ]
\end{equation}

\subsection{Aggregation mechanism based on Gaussian-DP}
PATE-TripleGAN employs input queries directed at teacher models for binary discrimination, with each teacher generating its binary classification results. Distinct aggregation strategies are adopted depending on the nature of the input data.

In the case of synthetic ``data-label" objects sampled from the generator, the aggregation mechanism introduces Gaussian noise independently to the votes for ``true" and ``false" categories, ultimately outputting the result with the highest noisy votes. 

To address the noisy aggregation in the generator, we employ Gaussian-DP as the privacy composition mechanism. To optimize the utilization of the privacy budget, we reference the Confidence-Aggregation method and set the threshold $T_e$ for teacher consensus\citep{papernot2018scalable}. The aggregation mechanism only returns conventional GNMax results under standard deviation $\sigma_2$ when the highest noisy vote with standard deviation $\sigma_1$ surpasses the threshold $T_e$. 

It is evident that augmenting the number of teachers broadens the acceptable range of the Gaussian noise standard deviation $\sigma$, thereby enhancing the privacy protection level of the model. However, this expansion comes with a trade-off as it reduces the amount of training data available for each teacher, resulting in a decline in the classification accuracy of individual teacher models. The noisy aggregation process can be mathematically expressed using the following formula:

\begin{equation}
    GNMax_\sigma(x_j,y_j) = \mathop{argmax}_{i\in[0,1]}[N_i(x_j,y_j)+ \mathcal{N} (0,\sigma^2) ]
\end{equation}

Conversely, for synthetic ``data-label" objects sampled from the classifier, the aggregation mechanism directly outputs the result with the highest votes for ``true" or ``false" categories, omitting the introduction of any noise. Given that the objects provided by the classifier inherently contain sensitive data, adding noise to the classifier's discernment results satisfies only privacy requirement ($II$) but falls short of meeting requirement ($I$). Privacy protection is achieved by introducing noise to the clipping gradients using the DPSGD method.

The aggregation's pseudo-code is shown in algorithm \ref{Aggregation}:
\begin{algorithm}
    \caption{Hybrid aggregation algorithm (HyGNMAx)}
    \label{Aggregation}
    
    \begin{algorithmic}[1]
        \STATE \textbf{Input:} {$(x,y)$, threshold $T_e$, noise parameters $\sigma_1$ and $\sigma_2$.}
        \STATE \textbf{Initialize:} {Teacher models $T=[T_1,...,T_k]$, Generator $G$, Classifier $C$.}
        \IF{ $(x,y)$ sampled from $C$}
            \STATE res $\gets {argmax}_{i\in [0,1]}\{ N_i(x,y) \}$
        \ELSIF{ $(x,y)$  sampled from $G$}
            \IF{ $max_i$\{ $N_i(x,y)$+$ \mathcal{N}(0,\sigma_1^2)\geq T_e$ \} }
                \STATE res $\gets argmax_{i \in [0,1]}$\{ $N_i(x,y)$+ $\mathcal{N}(0,\sigma_2^2)$ \}
            \ELSE
                \STATE break
            \ENDIF
        \ENDIF
        \STATE \textbf{Return} res
    \end{algorithmic}
\end{algorithm}

\subsection{Private GANs framework}
This article's primary contribution and innovation lie in introducing a hybrid gradient desensitization algorithm designed for discriminators. The central idea revolves around considering the privacy gradients of the discriminator as a combination of ``private gradients for the generator" and ``private gradients for the classifier." As a result, distinct gradient perturbation techniques are applied based on their respective data characteristics. This strategy can mitigate the risk of excessive gradient clipping and guarantee a more preservation of the original gradient information, thus improve the utility and convergence of the model.

The training process of PATE-TripleGAN is outlined in Algorithm \ref{Pate-TGAN}.

\begin{algorithm}
    \caption{PATE-TripleGAN}
    \label{Pate-TGAN}
    \begin{algorithmic}[1]
        \STATE \textbf{Input:} {Number of teachers $k$, Noise size for DPSGD $\sigma_0$, Clip parameter $R$, Iterations $n_s$, Classes set $M=[M_1,M_2,...,M_m]$, Batch Size $n_c$ and $n_g$, Gaussian-DP parameters $\mu_0 \;\&\;\delta $, Triple-GAN weight parameter $\alpha$.}
        
        \STATE \textbf{Initialize:} Parameters of Network: $\theta_C$, $\theta_G$, $\theta_S$, Teacher model set $T=[T_1,...,T_k]$, Consumed privacy budget $\hat{\mu}=0$, Disjoint labeled subsets $D_1$, $D_2$, ..., $D_k$ of size $\frac{|D|}{k}$, Unlabeled dataset $S_d$.
        
        \WHILE{$\hat{\mu}<\mu_0$}
            \STATE Training teacher models
            \STATE $T \gets trained \;[T_1,...,T_k]$
            \FOR{$\tau = 1$, ..., $n_s$}
                \STATE Sample $d_1$, ..., $d_{n_c} \;from\; S_d$
                \FOR{$i=1$, ..., $n_c$}
                    \STATE $r_{ci} \gets HyGNMax_\sigma(d_i,C(d_i))$
                \ENDFOR
                \STATE Sample $z_1, ..., z_{n_g} \sim P_Z,and \;y_1, ..., y_{n_g} \;from\; M$
                \FOR{ $j=1$, ..., $n_g$}
                    \STATE $r_{gj} \gets HyGNMax_\sigma( G(z_j|y_j), y_j)$
                \ENDFOR
                \STATE  Update the student model using DPSGD
                \STATE $Loss_c \gets \sum_{i=1}^{n_c} [r_{ci}logS(u_{ci})+\alpha(1-r_{ci})log(1-S(u_{ci}))]$
                \STATE $\nabla_{\theta s} \gets DPSGD(Loss_c, \sigma_0, -R, R)$
                \STATE  Update the student model using PATE
                \STATE $\nabla_{\theta s}$ - $\sum_{j=1}^{n_g} [r_{gj}logS(u_{gj})+(1-\alpha)(1-r_{gj})log(1-S(u_{gj})]$

            \ENDFOR    

            \STATE Update the Classifier
            \STATE Sample $d_1$, ..., $d_{n_c} \;from\; S_d$
            \STATE $\nabla_{\theta c}[\alpha\sum_{i=1}^{n_c}\log(1-S(d_i,C(d_i)) ]+ R_c+\alpha_p R_p$
            
            \STATE Update the generator
            \STATE Sample $z_1, ..., z_{n_g} \sim P_Z,and \;y_1, ..., y_{n_g} \;from\; M$
            \STATE $\nabla_{\theta g}[(1-\alpha)\sum_{i=1}^{n_g}\log(1-S(G(z_i|y_i),y_i)) ]$
            
            \STATE Update $\hat{\mu}$ according to Eqn. \ref{hatmu}
            \STATE $\hat{\mu}\gets \mu$
        \ENDWHILE
        \STATE \textbf{Output:} $G$
    \end{algorithmic}
\end{algorithm}

Specifically, ``data-label" pairs are sampled from both the generator and the classifier, denoted as $u_i$ ($i=1, 2, ...$). After the hybrid aggregation, each acquires the corresponding authenticity judgment, denoted as $r_i$. The loss function for the student discriminator is then expressed as a composite of  two components.

Introduce the measurement parameter between the generator and classifier as $\alpha$. For the sampled data from the classifier, the loss function is formulated as:
\begin{equation}
    Loss_1 = \sum_{j=1}^{n_c} r_{cj} \log S(u_{cj}) + \alpha(1-r_{cj}) \log(1-S(u_{cj}))
\end{equation}

Similarly, for the sampled data from the generator, the loss function is formulated as:
\begin{equation}
    Loss_2 = \sum_{i=1}^{n_g} r_{gi} \log S(u_{gi}) + (1-\alpha)(1-r_{gi}) \log(1-S(u_{gi}))
\end{equation}

The gradient, $Loss_2$, inherently adheres to the privacy specifications of $\mu-GDP$ due to the noisy aggregation. For the sensitive gradient, $Loss_1$, we employ the DPSGD method \citep{abadi2016deep}  with gradient clipping and the addition of gaussian noise for optimization. This ensures precise control over the sensitivity by limiting the gradient clipping value. Finally, in accordance with the post-processing nature of differential privacy, the only publicly disclosed generator strictly follows the privacy specifications of differential privacy.

Through a detailed privacy analysis, we demonstrate that PATE-TripleGAN achieves stringent protection in both data feature privacy and associate privacy of data-label correspondence.

\subsection{Privacy Analysis}
We analyze the privacy-preserving capability of the PATE-TripleGAN model through the following propositions, providing corresponding proof processes.

\textbf{Proposition 1.} The PATE mechanism aggregates the ``data-label" objects synthesized by the generator through noisy aggregation. One effective GNMax aggregation satisfies $\sqrt{2/\sigma_2^2}-GDP$.

\textbf{Proof.} Let $D_i$ denote the sensitive data set assigned to each teacher model. When changing one data point to obtain its neighboring dataset $D_i'$, in the worst case, this adjustment can lead to a change of value 1 for two categories (one category's voting result increases by 1, and the other decreases by 1). Therefore, adding noise to the voting results of the teacher models can be considered as a composite of  two Gaussian mechanisms with sensitivity $\Delta f = 1$ \citep{boenisch2022individualized}.

A GNMax aggregation outputs the category corresponding to the maximum $N_i(X) + \mathcal{N} (0,\sigma^2)$. The composite of  two Gaussian mechanisms with sensitivity $\Delta f = 1$ and standard deviation $\sigma_2$ still satisfies Gaussian-DP:
\begin{equation}
    G_\mu = G_{\mu_1}\otimes G_{\mu_2}, \; with\; \mu=\sqrt{\mu_1^2+\mu_2^2}
\end{equation}

Here $\mu_1=\mu_2=\sqrt{\Delta f/\sigma_2^2}$. Therefore, one effective GNMax aggregation confirms to $\sqrt{2/\sigma_2^2}-GDP$.

Let the student discriminator undergo T iterations. In each iteration, the classifier and generator contribute batches of ``data-label" pairs, with sizes denoted as $n_c$ and $n_g$ respectively, forming the training set. The discriminator optimizes gradients from the classifier through the DPSGD algorithm, with gradient clipping at a maximum upper limit of $R$ and Gaussian noise standard deviation set to $\delta*R$. Denote the size of the unlabeled dataset as $N_d$. We will employ these conditions to calculate the privacy budget consumed by the PATE-TripleGAN model.

\textbf{Proposition 2.}  PATE-TripleGAN adheres to the privacy protection standards outlined in Gaussian-DP. The privacy parameter is: $\mu = [(n_c^2/N_d^2 (e^{1/\delta^2}-1)+2n_g/\sigma_2^2 )T]^{1/2}$

\textbf{Proof.} In each iteration, the classifier samples a batch of $n_c$ data from $S_d$ with a probability of $p$. The maximum sensitivity of the gradient is restricted to $R$, and the Gaussian noise parameter is $\delta = R*\delta$, where $\delta$ stands for noise multiplier. The privacy mechanism under Gaussian-DP is defined as:
\begin{equation}
    f_c = pG_{1/\delta}+(1-p)Id
\end{equation}

The discriminator computes the noisy gradient for generator-synthesized data based on GNMax aggregation. Referring to proposition 1, a noisy aggregation satisfies $\sqrt{2/\sigma_2^2} - GDP$. Thus, after $n_g$ times GNMax, according to the composition theorem of Gaussian-DP\citep{dong2022gaussian}, we obtain:
\begin{equation}
    f_g = G_{\sqrt{2*n_g/\sigma_2^2}}= G_{\mu_g}
\end{equation}

In each iteration of the discriminator, the privacy mechanism can be conceptualized as a composite of the previously mentioned functions $f_c$ and $f_g$. Following T iterations, the maximum consumption of the privacy budget by PATE-TripleGAN is expressed as:
\begin{equation}
f = [(pG_{1/\delta}+(1-p)Id)\otimes G_{\mu_g}]^{\otimes T}
\end{equation}

Note that the Gaussian noise introduced at each instance is mutually independent. In the context of Gaussian-DP, the computation of privacy composition, as outlined in equation (3), reveals that the tensor product $\otimes$ operator adheres to both commutative and associative principles. Consequently, when evaluating the upper bound of privacy budget utilization in PATE-TripleGAN, Gaussian-DP's composition theorem allows for the repositioning of the discriminator's perturbed gradients for the classifier to the forefront and the generator's perturbed gradients to the rear. By leveraging the associative property of tensor product operators, the overall privacy budget consumption can be succinctly articulated as follows:
\begin{equation}
    f = [(pG_{1/\delta}+(1-p)Id]^{\otimes T} \otimes G_{\mu_g}^{\;\otimes T}
\end{equation}

For the first half term, \cite{bu2020deep} transformed it into the expression of $G-DP$ using the central limit theorem:
\begin{equation}
    [(pG_{1/\delta}+(1-p)Id]^{\otimes T} \rightarrow G_{\mu_c}
\end{equation}

where:
\begin{align}
       \mu_C &= p\sqrt{T(e^{1/\delta^2}-1)} \\
       &=\frac{n_c}{N_d}\sqrt{T(e^{1/\delta^2}-1)} 
\end{align}

In conclusion, in accordance with the Gaussian-DP composition theorem\citep{dong2022gaussian}, the amalgamation of several Gaussian mechanisms remains consistent with the Gaussian mechanism. Let $G_{\mu_G} \gets G_{\mu_g}^{\;\otimes T}$, we have:
\begin{equation}
  f = G_{\mu_C} \otimes G_{\mu_G} = G_{\mu}
\end{equation}

where:
\begin{align}
\label{hatmu}
        \notag
     \mu & = \sqrt{\mu_C^2+\mu_G^2}   \\
        & = \sqrt{\frac{n_c^2}{N_d^2}T(e^{1/\delta^2}-1)+\frac{2*n_g*T}{\sigma_2^2} }\\
        \notag
\end{align}

It becomes evident that the ultimate privacy budget remains unaffected by the dimensions of the labeled dataset ($S_l$). This outcome stems from the segmentation inherent in the PATE mechanism, which imposes rigorous constraints on the sensitivity of GNMax's noisy aggregation. Furthermore, in situations characterized by a substantial volume of unlabeled data and a comparatively limited amount of labeled data, the privacy budget expended by the hybrid gradient desensitization algorithm is notably diminished.

\textbf{Proposition 3.} The hybrid gradient desensitization algorithm can offer rigorous protection in both ``data feature privacy" and ``data-label correspondence privacy".

\textbf{Proof.} In the hybrid gradient desensitization algorithm, private gradients are classified into two categories: ``private gradients for the classifier" and ``private gradients for the generator."

For the former, the algorithm employs DPSGD to limit gradient sensitivity to a maximum value, denoted as R, and subsequently introduces Gaussian noise. This process aligns with the formulation under Gaussian-DP, represented as $f=pG_{1/\sigma}+(1-p)Id$. It is essential to note that the sensitivity of these gradients encompasses components from both the unlabeled dataset $S_d$ and the labeled dataset $S_l$. Therefore, the privacy protection provided by the DPSGD algorithm for gradients can be fundamentally broken down into a composite of three privacy mechanisms: $f_d \otimes f_{lx1} \otimes f_{ly1}$. Here, $f_d$ signifies the privacy protection mechanism for the unlabeled dataset, $f_{lx1}$ represents the privacy protection mechanism for the data portion of the labeled dataset, and $f_{ly1}$ accounts for the data-label correspondence of the labeled dataset.

Regarding the discriminator's gradients towards the generator, the algorithm utilizes PATE's GNMax aggregation for privacy protection. Importantly, the privacy component of these gradients only encompasses the sensitive voting results of the teacher models. Consequently, we can similarly break it down into a composite of  two privacy mechanisms: $f_{lx2} \otimes f_{ly2}$.

By consolidating similar privacy mechanisms, each iteration of the discriminator's gradient desensitization mechanism can be deconstructed into the composition of the following three mechanisms: $f_d \otimes f_{lx} \otimes f_{ly}$.

After T iterations, the privacy protection mechanism adheres to the $\mu-GDP$ specification outlined in Equation 22. According to the Gaussian-DP, the privacy levels of $f_d$, $f_{lx}$, and $f_{ly}$ will be no less stringent than $\mu$ \citep{dong2022gaussian}.

In conclusion, the hybrid gradient desensitization algorithm can afford robust protection in both ``data feature privacy" and ``data-label correspondence privacy".

\section{Experiment}
This section provides a summary evaluation of PATE-TripleGAN and the state-of-the-art model DPCGAN. The experimental conclusions are drawn based on two mainstream image datasets: MNIST and Fashion-MNIST.

\subsection{Experimental Setup}
To comprehensively assess the quality of synthetic data, we established a unified experimental procedure and standardized evaluation criteria. Initially, the training set is partitioned into two subsets, $S_l$ and $S_d$, representing labeled and unlabeled datasets respectively, based on the parameter `Percent'. Subsequently, PATE-TripleGAN and baseline models are trained on the standard datasets according to their respective configurations until the privacy budget limit is reached. This phase generates a series of checkpoints for the generators. Finally, the classifier models are trained on the private datasets synthesized by the trained generators and tested on real datasets, evaluating performance based on classification accuracy. Some fundamental parameter settings are as follows:

\begin{itemize}
    \item \textbf{Environment and Preparation}: The CPU is Intel(R) Xeon(R) Platinum 8255C, with one Nvidia RTX 2080 Ti (11GB) GPU. Python 3.8, Pytorch 2.0.0, and cuda 11.8 are employed. Both MNIST and FashionMNIST datasets are transformed into 32x32 image format.
    \item \textbf{DPCGAN}: Batch size is set to 128, with a learning rate $lr = 1e-4$. Privacy protection is implemented using Opacus's privacy engine, with a gradient clipping value (clip) of 1. The maximum epoch is set to 10 (when `Percent = 0.8', the generator's iteration count is 3750). The training process for each setting is repeated 5 times, with the best checkpoint in terms of generation quality selected for evaluation each time.
    \item \textbf{PATE-TripleGAN}: Batch size is also set to 128, with a learning rate $lr = 1e-4$. The maximum iteration counts for the generator are {500, 1000, 1500, 2000} respectively. Similar to DPCGAN, the training process for each setting is repeated 5 times, with the best-performing checkpoint selected for evaluation in each repetition.
\end{itemize}

We constructed a CNN image classifier called 'OurCNN'. It has 3 convolutional layers followed by ReLU activation and max-pooling. Additionally, there are two fully connected layers, and the last layer has 10 neurons for 10 classes. Additionally, we utilized a range of pre-defined Pytorch classifiers for evaluation, including Resnet18\citep{he2016deep}, Vgg16\citep{simonyan2014very}, DenseNet\citep{huang2017densely}, and Alexnet\citep{krizhevsky2012imagenet}.

\subsection{Experimental Results}
For each assessment parameter, we initially establish a set comprising 20,000 random noise instances paired with corresponding labels. This set remains fixed until a new evaluation metric is introduced. Subsequently, the generators access their respective checkpoints and  generate private datasets using random noise along with labels. 

The classifier models undergo 10 epochs of iterative training on these private datasets before proceeding to the evaluation phase. In each evaluation cycle, the classifier models handle the MNIST test dataset in batches of predefined sizes, capturing the count of correctly classified instances in each batch. Accuracy is determined by dividing the total number of correct classifications by the size of the test dataset. This evaluation process is iterated five times, and the average accuracy is computed to ascertain the evaluation outcome.

\begin{table*}
\begin{center}
\begin{tabular}{l c c c c} 
 \hline
                & \multicolumn{2}{c}{MNIST}     & \multicolumn{2}{c}{Fashion-MNIST}\\ 
 $\epsilon=10$  & DPCGAN    & PATE-TripleGAN    & DPCGAN    & PATE-TripleGAN  \\
 \hline
 OurCNN         & 0.5334    & \textbf{0.6073}   & 0.4915    & \textbf{0.5421} \\ 
 ResNet18       & 0.4190    & \textbf{0.4856}   & 0.4676    & \textbf{0.5194} \\ 
 Vgg16          & 0.6106    & \textbf{0.6280}   & 0.4770    & \textbf{0.5085} \\
 DenseNet       & 0.3223    & \textbf{0.3584}   & 0.3860    & \textbf{0.3908} \\
 AlexNet        & 0.6078    & \textbf{0.6901}   & 0.4952    & \textbf{0.5164} \\
 Average        & 0.4997    & \textbf{0.5539}   & 0.4635    & \textbf{0.4954} \\
 \hline
\end{tabular}
\caption{Generator Performance between DPCGAN and PATE-TripleGAN Trained on MNIST and Fashion-MNIST Datasets under $\epsilon=10$ and $Percent=0.8$.
}
\label{Accuracy_compare}
\end{center}
\end{table*}

\begin{figure*}[h]
  \centering
  \begin{subfigure}[b]{0.45\linewidth}
    \centering
    \includegraphics[width=\linewidth]{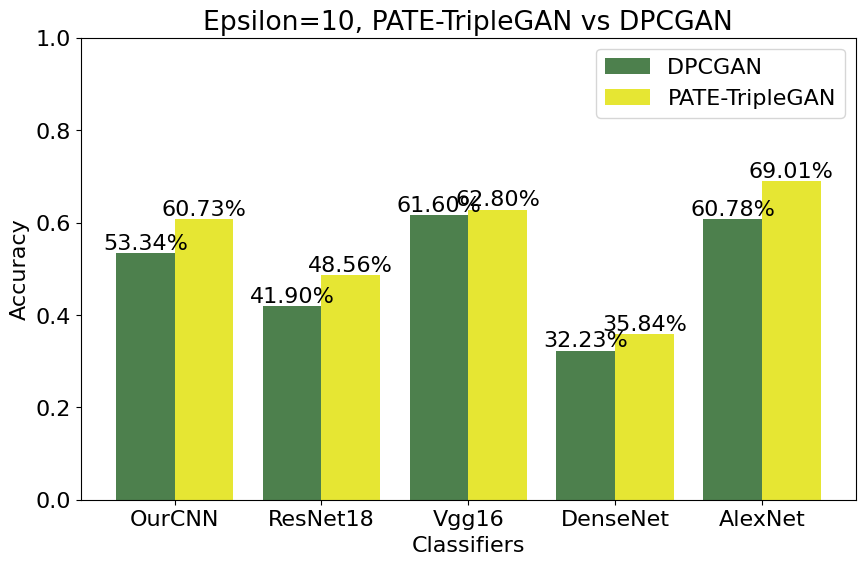}
    \caption{Rating outcomes for classifiers under conditions of $\epsilon=10$ and $Percent=0.8$.}
    \label{Epsilon=10}
  \end{subfigure}
  \hfill
  \begin{subfigure}[b]{0.45\linewidth}
    \centering
    \includegraphics[width=\linewidth]{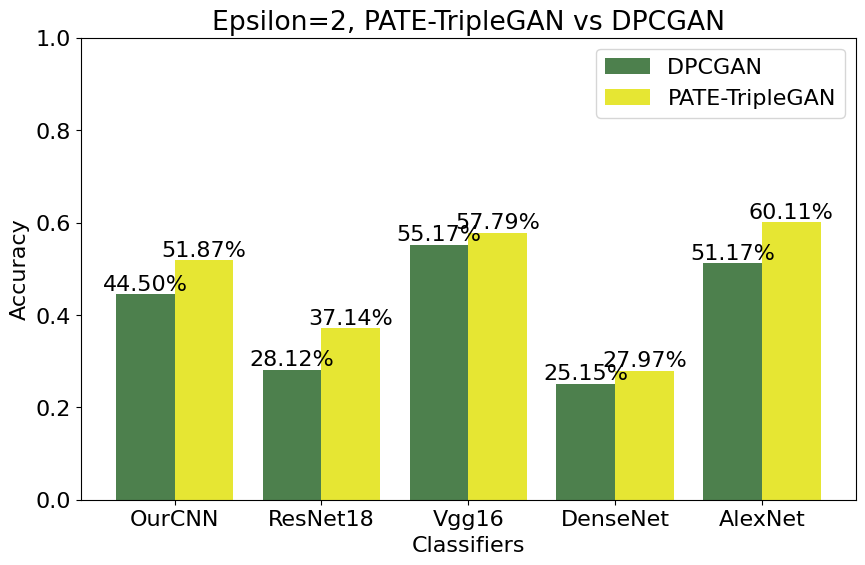}
    \caption{Rating outcomes for classifiers under conditions of $\epsilon=2$ and $Percent=0.8$.}
    \label{Epsilon=2}
  \end{subfigure}
  \caption{Comparative Histograms of Classifier Performance Ratings Utilizing Data Generated by PATE TripleGAN and DPCGAN Generators under Epsilon Values of 10 and 2.}
  \label{Different_epsilon}
\end{figure*}

\subsubsection{Performance of Generators under Different $\epsilon$}
In Table \ref{Accuracy_compare}, we first present the performance of generators from PATE-TripleGAN and DPCGAN trained on MNIST and Fashion-MNIST datasets under $\epsilon=10$ and $Percent=0.8$. The result indicates that classifiers trained on private datasets generated by PATE-TripleGAN exhibit stronger classification performance compared to those trained on datasets generated by DPCGAN. This underscores the capability of PATE-TripleGAN to produce higher-quality images compared to DPCGAN under consistent $\epsilon$ values.

We conducted an evaluation of the classification performance of diverse classifiers at an $\epsilon$ value of 2, depicted in Figure \ref{Different_epsilon}. In contrast to the situation where $\epsilon$ equals 10, there was a slight decline in the classification accuracy of each classifier. Nonetheless, PATE-TripleGAN demonstrated superior generation performance compared to DPCGAN. Moreover, when evaluated using classifiers such as AlexNet, the performance of PATE-TripleGAN at epsilon equals 2 closely resembled that of DPCGAN at epsilon equals 10.

\subsubsection{Hybrid Desensitization vs. Unified Clipping}
To examine the distinct impacts of both the hybrid gradient desensitization algorithm and the PATE mechanism within the model, we carried out two specific ablation experiments.

\begin{figure}[h]
    \centering
    \includegraphics[width=\linewidth]{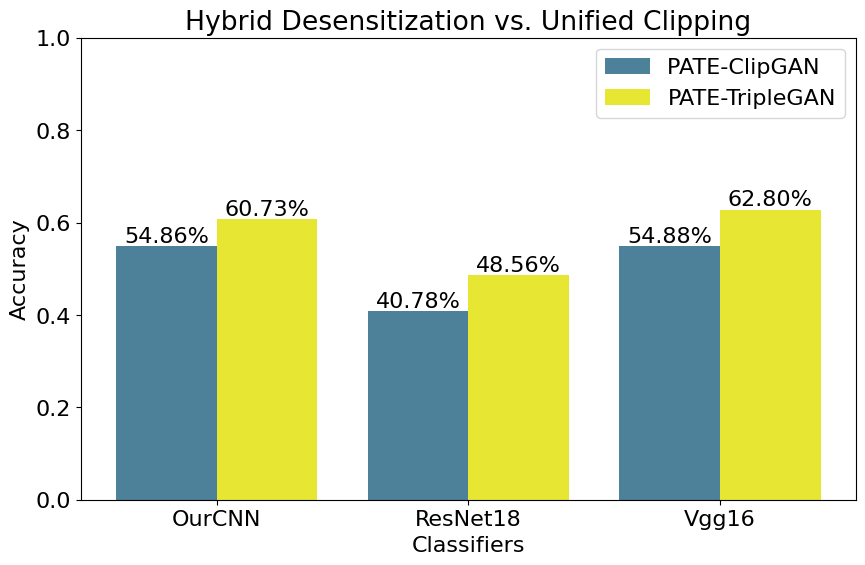}
    \caption{Due to the application of  hybrid gradient desensitization algorithm, PATE-TripleGAN can preserve the information of the original gradient better.}
    \label{hybrid_or_clip}
\end{figure}

In Figure \ref{hybrid_or_clip}, we introduce a novel model named PATE-ClipGAN, characterized by $\epsilon=10$ and $Percent=0.8$. In this model, we combine the gradients of the classifier and generator parts of the student discriminator and clip them together, followed by the introduction of Gaussian noise (parameters synchronized with PATE-TripleGAN), thereby ensuring privacy preservation. The notable performance degradation of PATE-ClipGAN's generator is evident, particularly when compared to DPCGAN in assessments involving ResNet18 and VGG16.

Notably, the PATE mechanism tends to consume a substantial portion of the privacy budget during computation, resulting in fewer overall iterations for the models (specifically, the generator's iteration counts in our experiment are {500, 1000, 1500, 2000}). Consequently, PATE-ClipGAN experiences a significant decline in performance due to its lower iteration count and the considerable noisy perturbations applied to the original gradients. Conversely, hybrid gradient desensitization, which effectively preserves the original gradients of the discriminator more, enhances the quality of adversarial learning, thereby exhibiting satisfactory performance even with reduced iterations.

\subsubsection{Performance under different percentages of data}

\begin{table*}
\begin{center}
\begin{tabular}{l c c c c c} 
 \hline
                & \multicolumn{3}{c}{Percent=0.8}     & \multicolumn{2}{c}{Percent=0.3}\\ 
 $\epsilon=10$  & CGAN    & DPCGAN  & PATE-TripleGAN  & DPCGAN    & PATE-TripleGAN  \\
 \hline
 OurCNN         & 0.9521  & 0.8721  & \textbf{0.9145} & 0.8218    & \textbf{0.8719} \\ 
 ResNet18       & 0.9978  & 0.8583  & \textbf{0.8980} & 0.7952    & \textbf{0.8621} \\ 
 Vgg16          & 0.9966  & 0.8226  & \textbf{0.8831} & 0.7668    & \textbf{0.8481} \\
 \hline
\end{tabular}
\caption{We tested the AUROC values of three classifiers at Percent=0.8 and Percent=0.3, respectively. PATE TripleGAN can demonstrate excellent performance even when the original dataset is small.
}
\label{AUROC}
\end{center}
\end{table*}

\begin{figure}[h]
  \centering
  \begin{subfigure}[b]{\linewidth}
    \centering
    \includegraphics[width=\linewidth]{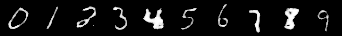}
    \caption{CGAN, Percent=0.8, not private}
    \label{x19}
  \end{subfigure}
  \begin{subfigure}[b]{\linewidth}
    \centering
    \includegraphics[width=\linewidth]{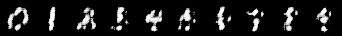}
    \caption{DPCGAN, Percent=0.8, $\epsilon=10$}
    \label{x1}
  \end{subfigure}
  \begin{subfigure}[b]{\linewidth}
    \centering
    \includegraphics[width=\linewidth]{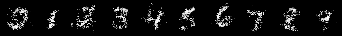}
    \caption{PATE-TripleGAN, Percent=0.8, $\epsilon$}
    \label{x500}
  \end{subfigure}
  
  \caption{In DPCGAN, the image contours exhibit reduced noise; however, distinguishing numbers 2, 3, 5, 6, 7, 8, and 9 proves challenging. PATE-TripleGAN yields images with noticeable noise; nonetheless, only numbers 2 and 8 present difficulty in recognition.}
  \label{result_pic}
\end{figure}

In Table \ref{AUROC}, we investigate the AUROC of DPCGAN and PATE-TripleGAN under $\epsilon=10$, $Percent=0.8$, and $Percent=0.5$. Consistent with prior observations \citep{ganev2021dp}, the incorporation of the PATE mechanism enhances the robustness of the generative model. Furthermore, owing to the inclusion of classifiers in adversarial training, PATE-TripleGAN can adeptly leverage the data from $S_d$ in comparison to DPCGAN, thus contributing to the improved performance of the generative model.

We present several images generated by CGAN, DPCGAN, and PATE-TripleGAN in Figure \ref{result_pic} under $\epsilon=10$ and $Percnet=0.8$. The image contours produced by DPCGAN are clearer; however, identifying numbers 2, 3, 5, 6, 7, 8, and 9 proves challenging. On the other hand, PATE-TripleGAN exhibits more image noise due to fewer iterations, yet only numbers 2 and 8 present difficulty in recognition.

\subsection{Discussion}
During the model debugging process, we conducted a detailed study on the hyperparameters of PATE-TripeGAN, including the number of teachers(k={50,100}), the gradient clipping value for the classifier (C=[1,5]), the noise multiplier ($mlp=[0.4,1]$), the number of iterations per round for teachers ($n_k=5$) and students ($n_s=\{5,10\}$), and the parameters of the PATE mechanism's Gaussian noise ($\sigma_2=[200,400]$).

Following are some insights we have gathered regarding PATE-TripleGAN throughout our experimental process:

\subsubsection{Iteration Count}
It is essential to note that the estimated iteration count for the generator in PATE-TripleGAN ranges from 1000 to 2000, significantly fewer than the approximate 4000 iterations of DPCGAN over 10 epochs. Indeed, the allocation of privacy budget to each iteration in PATE-TripleGAN is relatively higher compared to DPCGAN. This allocation affords us the flexibility to adopt less stringent gradient clipping values and noise multipliers(In this article, the value is 0.4) during DPSGD operations. As a result, more original gradient information can be preserved, contributing to improved model performance.

\subsubsection{Number of Teachers} 
We observed that the most effective performance of our model is attained when the quantity of teacher models falls within the range of 50 to 100, corroborating earlier findings \citep{jordon2018pate}. Despite the considerable Gaussian noise introduced during the voting process, each teacher benefits from a sufficient training dataset, enabling them to achieve precise discrimination and facilitating overall convergence. For instance, with specific parameterization, such as a batch size of 128, 100 teachers, and a standard deviation of 300 for the Gaussian noise ($\sigma_2$), the number of generator-synthesized data instances classified as real samples within each batch typically falls between 30 and 55. This contrasts with scenarios where half of the instances are classified as real and half as fake, which would make it difficult to distinguish between the two. Nevertheless, in the former case, the teacher model can effectively guide the learning of student discriminators.

\subsubsection{Classifier}
Given that the classifier is not publicly accessible, our focus lies solely on preserving the privacy of the classifier component within the discriminator's gradient. Consequently, during the training process, the classifier can draw data not only from the unlabeled set ($S_d$) but also from the labeled set ($S_l$). Under this circumstance, the parameter $N_d$ in the privacy analysis reflects the total dataset size (e.g., MNIST, comprising 60000 samples). It's noted that teacher models don't learn from the classifier's misclassifications, which may lead to a relatively high probability of misjudgment. To improve subsequent classification accuracy, all classification results can be initially considered as false, similar to TripleGAN, to prompt significant updates in the classifier model. After both the classifier and teacher model gradually converge, the PATE mechanism can then be employed.

\section{Conclusion}
This article presents a privacy-preserving training framework based on Gaussian-DP for PATE-TripleGAN, aiming to synthetically generate labeled data under the premise of privacy protection. The fundamental idea of PATE-TripleGAN involves introducing a classifier for pre-classifying unlabeled data, addressing the significant dependence of the DPCGAN model on labeled data. Additionally, a hybrid gradient desensitization algorithm is designed based on the the DPSGD algorithm and PATE mechanism to achieve model privacy protection. In contrast to the classic privacy protection approach of adding Gaussian noise on clipped gradients, PATE-TripleGAN can more effectively retain information from the original gradients, thereby enhancing the model's utility and convergence. Theoretical analysis and comprehensive experiments demonstrate that PATE-TripleGAN can preserve both ``data feature privacy" and ``data-label correspondence privacy", exhibiting stronger utility compared to DPCGAN in environments with low privacy budgets and limited labeled data. 

In future work, we will focus on exploring the feasibility of using PATE-TripleGAN for privacy-protected data synthesis in more complex scenarios, such as those with a larger number of categories and more intricate situations like cifar-100\citep{krizhevsky2009learning} and some tabular datasets.





\bibliographystyle{elsarticle-harv} 
\bibliography{example}






\end{document}